\documentclass[conference]{IEEEtran}

\usepackage{url}
\usepackage{cite}
\usepackage{amsmath,amssymb,amsfonts,bbm}
\usepackage{algorithmic}
\usepackage{graphicx}
\usepackage{textcomp}
\usepackage{xcolor}
\usepackage{subcaption,booktabs}

\def\BibTeX{{\rm B\kern-.05em{\sc i\kern-.025em b}\kern-.08em T\kern-.1667em\lower.7ex\hbox{E}\kern-.125emX}}

\begin{document}

\title{Identifying Surgical Instruments in Pedagogical Cataract Surgery Videos through an Optimized Aggregation Network
}

\author{\IEEEauthorblockN{Sanya Sinha}
\IEEEauthorblockA{\textit{Dept. of Surgery and Cancer} \\
\textit{Imperial College London}\\
London, United Kingdom\\
\url{s.sinha24@imperial.ac.uk}}
\and
\IEEEauthorblockN{Michal Balazia}
\IEEEauthorblockA{\textit{Team STARS} \\
\textit{INRIA d'Université Côte d'Azur}\\
Sophia Antipolis, France\\
\url{michal.balazia@inria.fr}}
\and
\IEEEauthorblockN{Francois Bremond}
\IEEEauthorblockA{\textit{Team STARS} \\
\textit{INRIA d'Université Côte d'Azur}\\
Sophia Antipolis, France\\
\url{francois.bremond@inria.fr}}
}

\maketitle

\begin{abstract}
Instructional cataract surgery videos are crucial for ophthalmologists and trainees to observe surgical details repeatedly. This paper presents a deep learning model for real-time identification of surgical instruments in these videos, using a custom dataset scraped from open-access sources. Inspired by the architecture of YOLOV9, the model employs a Programmable Gradient Information (PGI) mechanism and a novel Generally-Optimized Efficient Layer Aggregation Network (Go-ELAN) to address the information bottleneck problem, enhancing Minimum Average Precision (mAP) at higher Non-Maximum Suppression Intersection over Union (NMS IoU) scores. The Go-ELAN YOLOV9 model, evaluated against YOLO v5, v7, v8, v9 vanilla, Laptool and DETR, achieves a superior mAP of 73.74 at IoU 0.5 on a dataset of 615 images with 10 instrument classes, demonstrating the effectiveness of the proposed model.
\end{abstract}


\section{Introduction}

Pedagogical surgical videos benefit medical students by allowing them to explore surgical processes~\cite{fujii2022}. These videos are especially useful for minimally-invasive outpatient procedures, like cataract surgeries, by demonstrating the steps for trainees~\cite{nwoye2019a}. High-quality instructional videos let ophthalmologists and trainees repeatedly observe surgical details. Detecting tools used in these procedures helps estimate the type and position of surgical equipment. However, real-time tool detection is challenging due to the lack of annotated data. Open-access videos often contain patient faces, personal information, and are of poor quality, filmed on head-mounted cameras. This highlights the need for a comprehensive, annotated dataset with high-quality images.
Since a while, several object tracking technologies have been used to gauge the position and the presence of certain surgical equipment. Kranzfelder et al.~\cite{kranzfelder2013} leveraged radio frequency identification (RFID) technology to identify surgical equipment in minimally invasive real-time surgeries. Hasse et al.~\cite{haase2013} suggested a time-of-flight and RGB color information endoscopy-based tracking. However, both these traditional systems require added operational knowledge and costs. For example, RFID systems require specialized tags and readers, which can add to the overall cost, and endoscopy equipment tends to be expensive and require specialized training for operation. Their overall invasiveness renders them useless for minimally-invasive ophthalmic surgeries like cataract.

\begin{figure*}[t]
\vspace{-10pt}
\centering
\includegraphics[width=\textwidth]{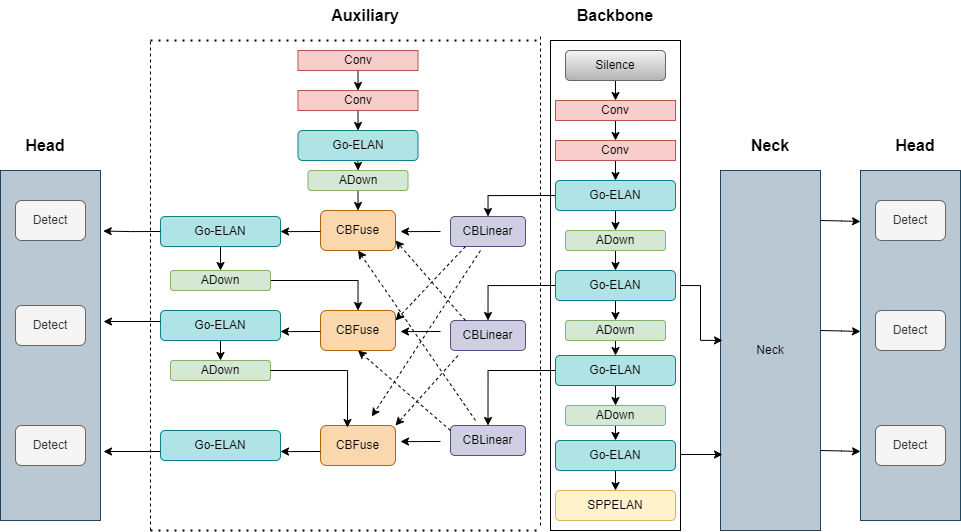}
\caption{Go-ELAN YOLOV9 Complete Architecture. The Auxiliary block works on the Programmable Gradient Information (PGI) concept by creating an auxiliary reverse branch for enabling reliable gradient calculation by avoiding potential semantic loss. The GELAN block in the backbone feature extractor is replaced by the Go-ELAN block proposed in this paper. The Spatial Pyramid Pooling block SPPELAN removes the fixed size limitation of the backbone. The ADown block downsamples the generated feature maps to target sizes. the CBLinear blocks extract higher level features from the images, and the CBFuse block fuses these extracted features. The Neck combines the acquired features and the Head predicts the final bounding bound outputs with their respective probabilities.}
\label{fig:yolov9}
\vspace{-10pt}
\end{figure*}

\section{Related Work}

With the advent of AI in surgery and robot-assisted intervention techniques, deep learning has positively impacted object detection. Numerous researchers have contributed to the advancement of AI-driven surgical tool detection methodologies. For instance, Twinanda et al.~\cite{twinanda2017} introduced the baseline model EndoNet, which performs tool presence detection and phase recognition tasks simultaneously. Sarikaya et al.~\cite{sarikaya2017} utilized a region proposal network and a multi-modal two-stream convolutional network for tool detection. Kurmann et al.~\cite{kurmann2017} proposed a U-Net architecture-based model that jointly performs tool detection and 2D pose estimation. Additionally, Jin et al.~\cite{jin2018} achieved high detection accuracy using region-based CNNs (R-CNNs). Hajj et al.~\cite{alhajj2019} applied a CNN-RNN model to detect tools in surgical videos, employing a boosting mechanism instead of end-to-end training. Nwoye et al.~\cite{nwoye2019b} developed an end-to-end approach composed of CNN-convolutional LSTM (ConvLSTM) neural networks for tool presence detection and tracking using tool binary labels. Wang et al.~\cite{wang2017} proposed a method that combines 3D CNNs and graph convolutional networks (GCNs) for tool presence detection, considering the relationship between tools. Jin et al.~\cite{jin2020} presented a multi-task recurrent convolutional network with correlation loss (MTRCNet-CL) for tool presence detection and surgical phase recognition. Even in the domain of open surgery, AI has been widely used to solve detection problems. Shimizu et al.~\cite{shimizu2020} introduced an innovative surgical recording system that employs multiple cameras placed on a surgical platform. This setup leverages computer vision-based techniques for region segmentation and recognition, facilitating automatic camera selection to capture optimal view and mitigate occlusion issues, resulting in a unified video output. Based on this work, Hachiuma et al.~\cite{hachiuma2020} improved the camera selection algorithm using CNN, aiming to further refine the selection process. Yoshida et al.~\cite{yoshida2021} tackled the challenge of estimating incision scenes in lengthy open surgery videos by analyzing factors such as gaze speed, hand movements, the number of hands involved, and background dynamics within egocentric surgical footage.

While all these methods successfully harnessed the available data to detect surgical features, there is still not satisfactory progress made in the field of AI-assisted pedagogical surgical video analysis to support medical personnel-in training. Choi et al.~\cite{choi2017} proposed the use of YOLO to detect crucial surgical features in laparoscopic surgeries. Through this paper, we aim to present a unique, optimized approach inspired by YOLOV9~\cite{wang2024} to detect 10 classes of surgical instruments used for cataract surgery through a dataset created by scraping open-access pedagogical videos on a frame-by-frame basis. We strive to create a light-weight object detection model that would supersede the performance of YOLOV9 while leveraging the same block components.

\section{Methodology}

In deep neural network models, there is a constant risk of information loss during data traversal across network layers. This is characterized through the information bottleneck problem during the feed-forward process. The information bottleneck problem can undermine network performance and reduce overall model efficiency. Consequently, several methods evolved to retain information even across network depths to overcome the information bottleneck challenges. Reversible architectures~\cite{cai2023} address this issue by enabling the computation of activations in a reversible manner, allowing intermediate activations to be recomputed from the output during back-propagation without the need to store them explicitly. This approach relies on bijective transformations to significantly reduce memory consumption during training, thus enabling the training of deeper networks. Masked modeling~\cite{chen2022} is another method to overcome the information bottleneck problem. It relies primarily on the loss of reconstruction and employs an implicit method to enhance the extraction of features while preserving the input information. However, the loss of reconstruction of mask models often interferes with the loss of the target, reducing the computational accuracy of the model. Deep supervision models rely on features that have not lost significant information to establish feature-to-target maps for information traversal across deeper network layers. However, if the shallow features have lost a major share of information, they would hamper the learning performance. 

In the YOLOV9 model, a novel Programmable Gradient Information mechanism is launched which facilitates the creation of reliable gradients through an auxiliary reversible branch. Thus, the gradient information is programmed at different semantic levels to achieve the best performance. The use of an auxiliary branch reduces the net cost of the model, and the calculated semantic loss does not interfere with the target loss, unlike mask modeling. In our proposed model, the Programmable Gradient Information interface is amalgamated with an optimized version of a Generalized ELAN (GELAN) architecture to aid the development of lightweight and high-performing object detection models.

\subsection{Programmable Gradient Information}

Programmable Gradient Information (PGI) is a concept central to enhancing the training process of machine learning models by providing the ability to manipulate gradients. Gradients are typically computed automatically based on the loss function and propagated backward through the network via techniques like backpropagation. However, in certain scenarios, it becomes beneficial to programmatically modify these gradients to achieve specific objectives or address challenges encountered during training. PGI mainly includes three components, namely (1) main branch, (2) auxiliary reversible branch, and (3) multilevel auxiliary information, where the main branch performs inference, the auxiliary branch deals with the information bottleneck, and the multi-level auxiliary branch manages error accumulation due to deep supervision. The proposed model updates information in the main inference branch through the gradients obtained from the reversible auxiliary branch. This design is effective on both deep and shallow networks.

\subsection{Optimized GELAN: Go-ELAN YOLOV9}

Generalized Efficient Layer Aggregation Network (GELAN) is an amalgamation of CSPNet~\cite{wang2020}, used by YOLOV8 and ELAN~\cite{redmon2018}, used by YOLOV3. The GELAN model's backbone is structured to extract hierarchical features from input images through a series of convolution operations and specialized blocks. To detect surgical instruments in videos, it is important to have a greater mAP than an F1 score. Since the vanilla version of YOLOV9 is based on GELAN, the experimental results show that it has an unsatisfactory mAP-to-F1 ratio. To address this problem, our work deals with developing a modified version of YOLOV9 by optimizing the GELAN architecture. It starts with a convolutional downsampling step (P1/2), employing a 3x3 kernel with 512 filters and a stride of 2. This is followed by a subsequent downsampling layer (P2/4) with similar parameters but employing 512 filters instead of 128. The backbone then incorporates ELAN-1 and ELAN-2 blocks. After each downsampling step, the model progresses to average-convolution downsampling layers, such as at P3/8 and P4/16, which further refine the feature representation by increasing receptive fields and feature map dimensions. ELAN-2 blocks are applied recurrently to facilitate feature extraction and information fusion on different scales, maintaining consistency in the hierarchical feature learning process. Finally, the backbone concludes with a last average-convolution downsampling step (P5/32), preparing the feature maps for further processing in the model's head. The backbone leads to the model's neck, which functions as a feature aggregator. The aggregated image features are finally passed into the model's head for prediction. The complete proposed model architecture is in Figure~\ref{fig:yolov9}.

\begin{figure}[ht]
\vspace{-5pt}
\centering
\includegraphics[width=\linewidth]{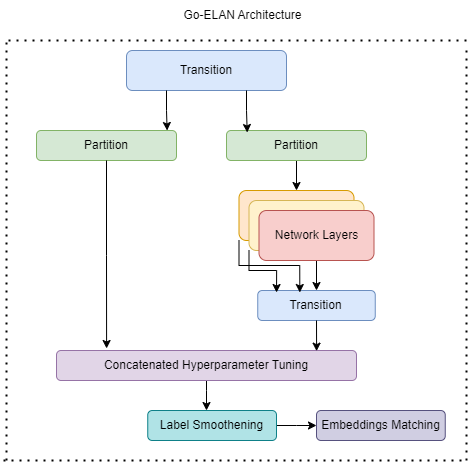}
\caption{Go-ELAN Architecture: Size of downsampling filters increases from 128 in GELAN to 512 in Go-ELAN to accommodate greater spatial context. A label smoothener is added in the loss computer to spread out the probability mass.}
\label{fig:Go-ELAN-YOLOV9}
\end{figure}

Likewise, changes are made to the model’s detector head while producing final predictions. The regularization parameters of this Generally-optimized ELAN (Go-ELAN) are fine-tuned to 0.01~\cite{mensbrugghe2023} and a label-smoothening block with smoothening coefficient of 0.1 is introduced to the loss computing framework to ensure soft targets by spreading out the probability mass from the true label to other incorrect labels, as in Figure~\ref{fig:Go-ELAN-YOLOV9}. The Go-ELAN YOLOV9 modification in the YOLOV9 architecture significantly improves the model’s mAP at higher NMS IoU and ensures a better precision-to-recall trade-off. We express the mutual information involving the Go-ELAN YOLOV9 function with its parameters $\phi$ and $\psi$ as
\[
I(X,X) = I\left(X,g_{\phi}(X)\right) = I\left(X,\upsilon_{\psi}\left(g_{\phi}(X)\right)\right)
\]
where $I$ denotes mutual information, $g$ is the Go-ELAN YOLOV9 function, and $\phi$ and $\psi$ are respective parameters.

\section{Experiments and Results}

\begin{table*}[t]
\vspace{-10pt}
\centering
\caption{Performance metrics of various models.}
\vspace{-5pt}
\begin{tabular}{l|rrrrr}
\toprule
\textbf{Model Names} & \textbf{Average Precision} & \textbf{Average Recall} & \textbf{mAP(50\%)} & \textbf{mAP(95\%)} & \textbf{F1 Score} \\
\midrule
YOLOV5~\cite{zhu2021} & 0.613 & 0.652 & 0.712 & 0.514 & 0.631 \\
YOLOV7~\cite{wang2020} & 0.772 & 0.518 & 0.675 & 0.475 & 0.620 \\
YOLOV9 vanilla~\cite{wang2024} & 0.547 & \textbf{0.743} & 0.663 & 0.457 & 0.630 \\
YOLOV8~\cite{wang2023} & 0.575 & 0.526 & 0.564 & 0.409 & 0.549 \\
DETR~\cite{carion2020} & 0.570 & 0.453 & 0.245 & 0.225 & 0.504 \\
Laptool~\cite{namazi2022} & 0.487 & 0.560 & 0.620 & 0.495 & 0.520 \\
YOLOV9 Go-ELAN (proposed) & \textbf{0.859} & 0.598 & \textbf{0.723} & \textbf{0.525} & \textbf{0.705} \\
\bottomrule
\end{tabular}
\label{tab:model_performance}
\vspace{-10pt}
\end{table*}

\subsection{Dataset}

For this project, we created a custom cataract surgery dataset by scraping open-access instructional surgical videos. We referred to the open-access surgical videos~\cite{devgan2020, megur2021,shelby2019,cybersight2020} and generated a dataset by extracting each frame of information from the videos. Through the oral description of the surgical tools provided in the videos, we successfully annotated the surgical tools in the dataset using Roboflow.
The instruments in the videos were broadly classified into 10 classes: cannula, crescent blade, fixation ring, forceps, hook, keratome, needle, phacoprobe, speculum, and instruments. The ‘instruments’ class was provided for annotating any unspecified instruments whose labels could not be found. In the dataset, we also have a class labeled ‘speculum’ which contains image instances of Lancaster speculum as opposed to the Baraquer speculum used traditionally for cataract surgery. While there was a mention of the Lancaster speculum in the instructional voice-over, there was no video presence of the said speculum. Hence, the confusion matrix for the proposed method contains a mention of the Lancaster speculum class, but has no present instances. We were able to generate a dataset of 247 images through video analysis. However, extracting frames from videos compromises the image quality. Hence, we performed general image augmentation on the dataset through techniques like random cropping, horizontal, and vertical flipping to increase the number of data samples. Moreover, since low-light and poor-quality images cannot be used in the training framework, we have performed simple contrast-limited adaptive histogram equalization on the images. After the data augmentation techniques, we have a final dataset size of 615 images with 552 training images, 42 validation images, and 21 test images respectively. Since we have limited data samples, we have restricted the train:test:val ratio to 0.9:0.07:0.03.

\subsection{Experimental Results}

The Go-ELAN YOLOV9 framework was trained on a NVIDIA T4 GPU for 20 epochs. The SGD optimizer with parameters initial learning rate $0.01$, final learning rate $0.01$, momentum $0.937$, weight decay $0.0005$, warmup epochs $3.0$ and warmup momentum $0.8$. Blur $0.01$ and CLAHE~\cite{sinha2023} were used as albumentations. Data Augmentations involved included scale $0.9$, shear $0.0$, perspective $0.0$, lateral flip $0.5$, mosaic $1.0$ and mixup $0.15$. Downstreaming the modified YOLOV9 on our dataset included fine-tuning hyperparameters of batch size $8$, image size $640\times640$ and close mosaic $15$. The model has 50.9 million parameters, similar to that of YOLOV9, for better fitting of complex data. It also requires 237 GFLOPS for performing calculations, an indicator of higher computational cost. 

To prove the efficiency of our model, we have compared its performance with five other state-of-the-art models. YOLOV5, YOLOV7, YOLOV8, YOLOV9-vanilla, and DETR, and a surgical instrument detection model 'Laptool'~\cite{namazi2022}. The overall performance is compared according to the class-average F1 score, and Minimum Average Precision at an IoU score of $50\%$ and $95\%$. Table~\ref{tab:model_performance} illustrates the performance of various cutting-edge models in the context of surgical instrument detection. In particular, Go-ELAN YOLOV9 emerges as a top performer, boasting the highest Average Precision (AP) score of $0.829$ among all models. The performance of the model is validated through visual examinations, a confusion-matrix, quantitative metrics, and a precision-recall curve. 

This signifies its exceptional accuracy in pinpointing surgical instruments within medical images. Additionally, Table~\ref{tab:model_performance} shows that Go-ELAN YOLOV9 achieves a commendable mean Average Precision (mAP) of $0.723$ at 50\% Intersection over Union (IoU), demonstrating its robustness in accurately detecting instruments across different scenes. Even at the stringent 95\% IoU threshold, Go-ELAN YOLOV9 maintains a competitive mAP of $0.525$, indicating its ability to precisely identify instruments with minimal overlap. These impressive metrics collectively highlight Go-ELAN YOLOV9 as a superior choice for surgical instrument detection tasks, offering a compelling balance between precision, recall, and overall performance. Compared to Go-ELAN YOLOV9, the other models show various degrees of performance in surgical instrument detection. YOLOv7 emerges as a strong competitor, with an AP of $0.772$ suggesting great accuracy in detecting surgical equipment. However, it falls short of Go-ELAN YOLOV9's AP rating. Similarly, while YOLOv5 has a decent AP of $0.613$, it fails to compete with Go-ELAN YOLOV9's precision. Although YOLOv8 and YOLOv9 have good AP ratings, they lack precision, recall, and mAP compared to Go-ELAN YOLOV9. DETR, and Laptool, while competitive in terms of F1 Score, have lower mAP values, indicating challenges in reliably recognizing surgical equipment. Overall, while all models show promise in surgical instrument recognition, the Go-ELAN YOLOV9 stands out.

\subsection{Qualitative Results}

\begin{figure*}[t]
\vspace{-10pt}
\centering
\includegraphics[width=0.75\textwidth]{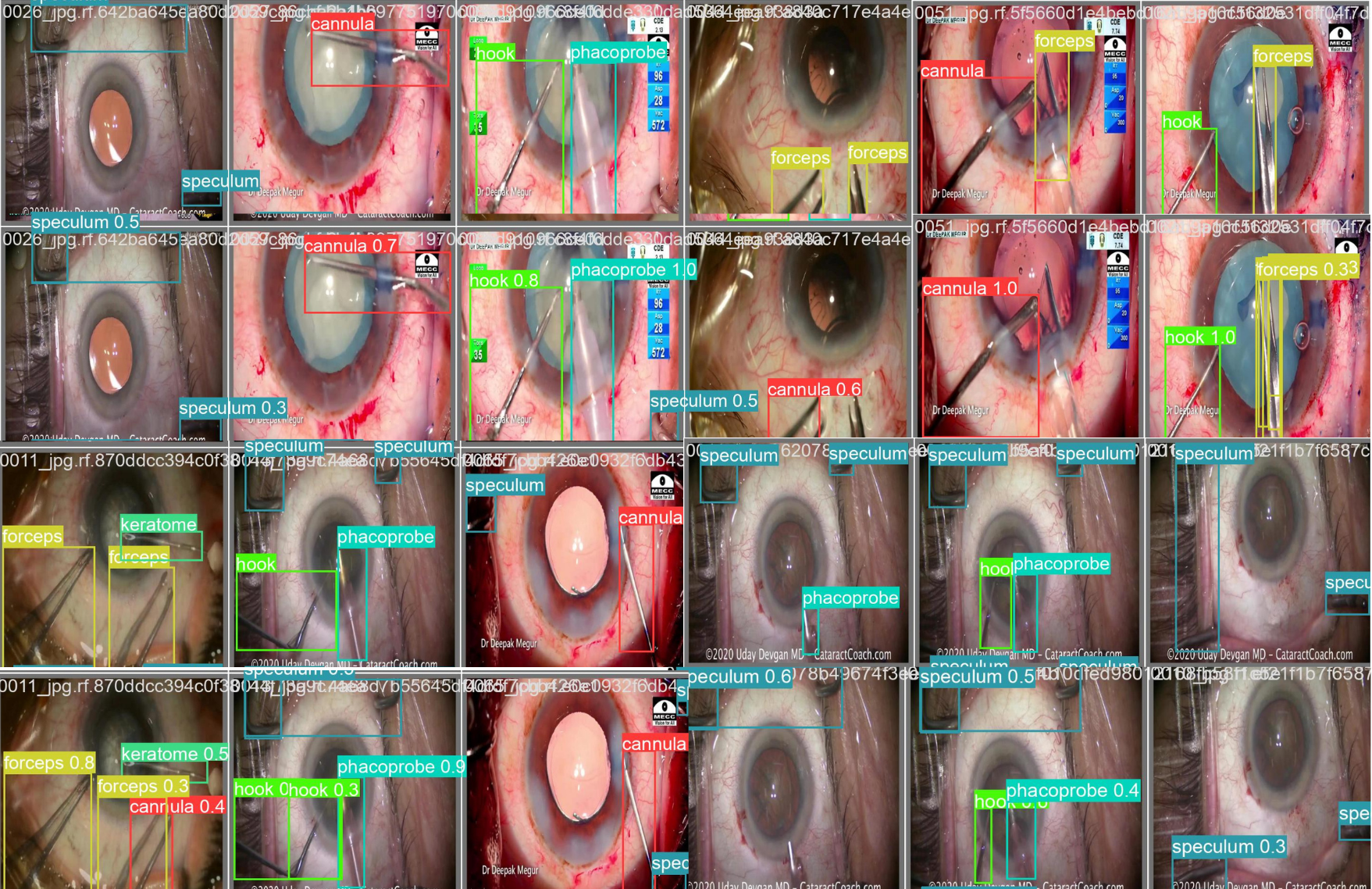}
\vspace{-3pt}
\caption{Qualitative Examination of Model Performance. Rows 1 and 3 are labels while 2 and 4 are respective predictions.}
\label{fig:outputs}
\vspace{-10pt}
\end{figure*}

To illustrate the visual and qualitative superiority of our model, we have compared 12 ground-truth images with their respective model predictions in Figure~\ref{fig:outputs}.

\begin{figure*}[t]
\centering
\begin{subfigure}[t]{0.49\textwidth}
\centering
\includegraphics[height=7cm]{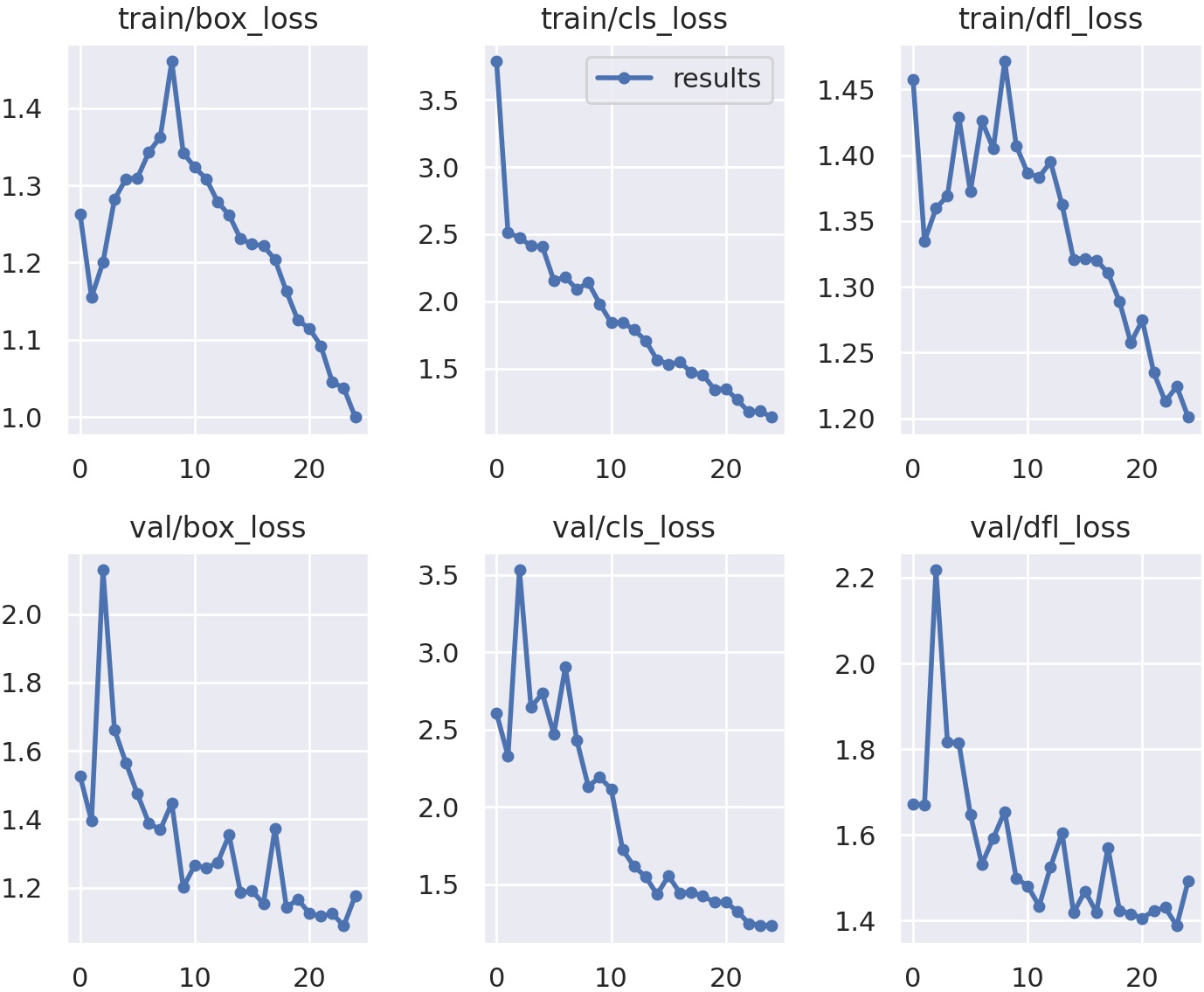}
\vspace{-3pt}
\caption{Quantitative Metrics of Model.}
\label{fig:results}
\end{subfigure}
\hfill
\begin{subfigure}[t]{0.50\textwidth}
\centering
\includegraphics[height=7cm]{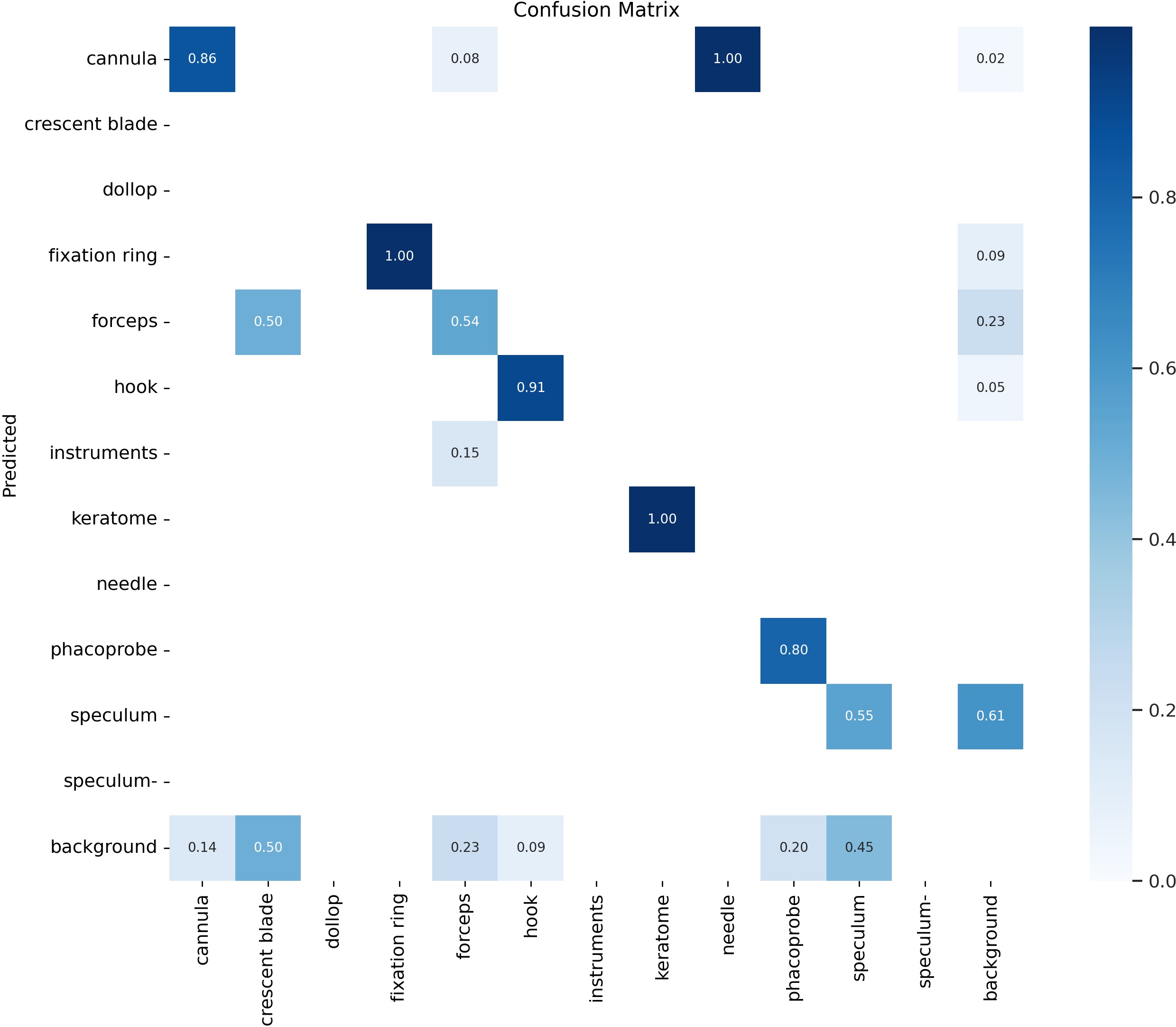}
\vspace{-3pt}
\caption{Confusion Matrix of Proposed Model.}
\label{fig:confusion matrix}
\end{subfigure}
\vspace{-3pt}
\caption{Qualitative and Quantitative Evaluation of the Model.}
\vspace{-14pt}
\end{figure*}

The displayed images show a total of 29 instruments (repetitions included). Out of the total 29 instruments present, the model correctly identified 23. There were a few instruments not annotated in the ground truth due to visibility constraints. However, the model was even successful in identifying and detecting them! This highlights the efficiency of the model in real-time surgical video analysis. The quantitative metrics present in Figure~\ref{fig:results} showcase the receding values of box loss, class loss, and distributed focal loss after each training and validation epoch. This shows a positive trend, as the model is gradually able to fit more complex data better. The performance of the model can also be evaluated through the confusion matrix in Figure~\ref{fig:confusion matrix}, which provides a detailed analysis of the model and a breakdown of its performance on different instrument classes. As seen, the model successfully identified all instances of the fixation ring and the keratome, and almost all instances of  surgical instruments including the cannula (0.86), the hook (0.91), and the phacoprobe (0.80), which appear to be quite similar visually. The model, however, posed a notable confusion between some classes, with forceps and speculum often misclassified as the background (instrument-free) with a score of $0.23$ and $0.45$. Instruments such as the dollop were also not identified at all. However, the dollop only had one visible instance across the dataset, so the model’s confusion is justified. Overall, instruments with a high frequency of occurrence in the video frames were correctly identified in almost all the cases.

To evaluate the performance of the model, we used the following two loss functions. Focal loss is defined as
\[
\mathcal{L}_{\text{focal}} = -\alpha (1 - \hat{p}_t)^\gamma \log(\hat{p}_t)
\]
where $\alpha$ is a balancing factor, $\gamma$ is a focusing parameter, and $\hat{p}_t$ is the predicted probability of the target classes.
It is, therefore, observed that since the model obtains lower values for the losses after each epoch, the model has optimum performance. Bounding box loss function is defined as
\[
\begin{split}
\mathcal{L}_{\text{box}} & = \lambda_{\text{coord}} \sum_{i=1}^{S^2} \sum_{j=1}^{B} \mathbbm{1}_{ij}^{\text{obj}} \left[ (x_i - \hat{x}_i)^2 + (y_i - \hat{y}_i)^2 \right] + \\
& + \lambda_{\text{coord}} \sum_{i=1}^{S^2} \sum_{j=1}^{B} \mathbbm{1}_{ij}^{\text{obj}} \left[ (\sqrt{w_i} - \sqrt{\hat{w}_i})^2 + (\sqrt{h_i} - \sqrt{\hat{h}_i})^2 \right]
\end{split}
\]
where $\lambda_{\text{coord}}$ is a weighting factor, $S$ is the grid size, $B$ is the number of bounding boxes, $\mathbbm{1}_{ij}^{\text{obj}}$ is an indicator function that denotes if object $j$ appears in cell $i$, and $x_i, y_i, w_i, h_i$ and their hat counterparts are the ground truth and predicted bounding box parameters respectively.

\section{Conclusion and Future Work}

We have developed a novel dataset of cataract surgical instruments by scraping information frames from open access cataract surgery videos. The dataset is available at RoboFlow with public access: https://universe.roboflow.com/sanya-vuzrm/cataract-7smf8/dataset/3. We have also developed a novel model influenced by the recent YOLOV9 architecture through the modification of the GELAN architectural block into the optimized, Go-ELAN YOLOV9 version. Our model returned an F1 score of $70.5\%$, and an mAP ($50\%$) of $72.3\%$, which exceeded the performance of other state-of-the-art object detection models. For the future, we plan to expand our work towards improving the average recall of the model to benefit surgeons in training with real-time instrument tracking and identification. We also intend to develop a live captioning system to highlight the role of each instrument in the surgical procedure in real-time by providing a textual response. The utility for this model could be extended beyond cataract surgery to surgical planning, robotic assistance, and patient monitoring to name a few.

\end{document}